\pdfoutput=1

\documentclass[11pt]{article}

\usepackage[]{acl}

\usepackage{times}
\usepackage{latexsym}
\usepackage{graphicx}
\usepackage{enumitem}
\usepackage{multirow}
\usepackage[ruled,linesnumbered]{algorithm2e}

\usepackage[T1]{fontenc}

\usepackage[utf8]{inputenc}

\usepackage{microtype}
\usepackage{smile}
\usepackage{graphics}
\usepackage{amsmath}

\newcommand*{\Scale}[2][4]{\scalebox{#1}{$#2$}} 

\title{Concept2Box: Joint Geometric Embeddings for Learning Two-View Knowledge Graphs}

\author{Zijie Huang$^{1}$\thanks{~~Part of work was done during internship at Amazon.}, Daheng Wang$^{2}$, Binxuan Huang$^{2}$, Chenwei Zhang$^{2}$, Jingbo Shang$^{4}$,\AND  Yan Liang$^{2}$, Zhengyang Wang$^{2}$, Xian Li$^{2}$, Christos Faloutsos$^{3}$, Yizhou Sun$^{1}$, Wei Wang$^{1}$\\
$^{1}$University of California, Los Angeles, CA, USA $^{2}$Amazon.com Inc, CA, USA\\
$^{3}$Carnegie Mellon University, PA, USA $^{4}$University of California, San Diego, CA, USA\\
\texttt{\{zijiehuang,yzsun,weiwang\}@cs.ucla.edu},\\
\texttt{\{dahenwan,binxuan,cwzhang,ynliang,zhengywa,xianlee\}@amazon.com},\\
\texttt{jshang@ucsd.edu,christos@cs.cmu.edu}\\
}

\begin{document}
\maketitle
\begin{abstract}

Knowledge graph embeddings (KGE) have been extensively studied to embed large-scale relational data for many real-world applications. 
Existing methods have long ignored the fact many KGs contain two fundamentally different views: high-level ontology-view concepts and fine-grained instance-view entities. They usually embed all nodes as vectors in one latent space. 
However, a single geometric representation fails to capture the structural differences between two views and lacks probabilistic semantics towards concepts' granularity. 
We propose Concept2Box, a novel approach that jointly embeds the two views of a KG using dual geometric representations. We model concepts with box embeddings, which learn the hierarchy structure and complex relations such as overlap and disjoint among them. Box volumes can be interpreted as concepts' granularity. 
Different from concepts, we model entities as vectors. To bridge the gap between concept box embeddings and entity vector embeddings, we propose a novel vector-to-box distance metric and learn both embeddings jointly.
Experiments on both the public DBpedia KG and a newly-created industrial KG showed the effectiveness of Concept2Box.

\end{abstract}

\begin{figure}[t]
	\centering
	\includegraphics[width=\linewidth]{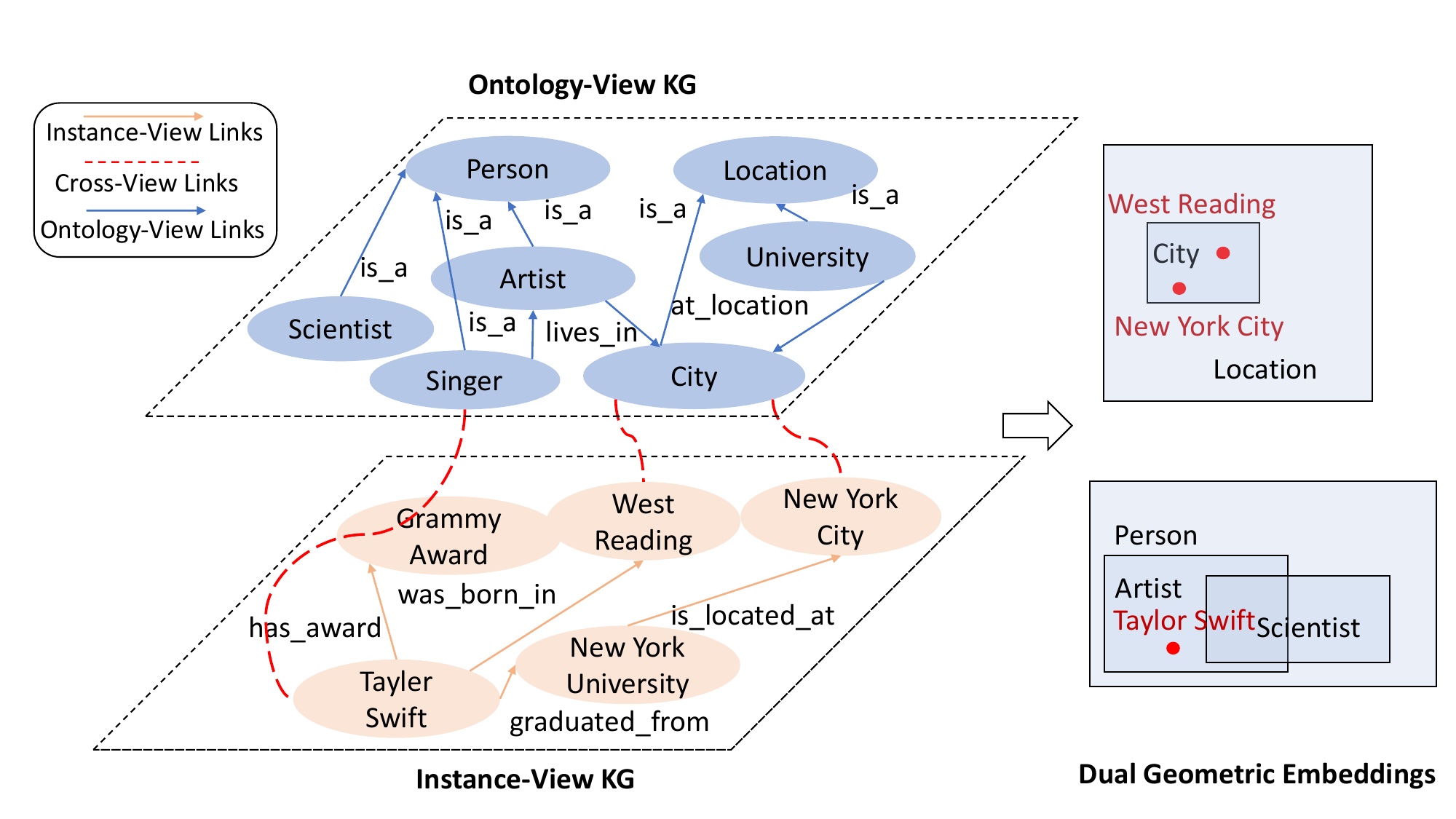}
	\caption{A two-view KG consisting of (1) an ontology-view KG with high-level concepts and meta-relations, (2) an instance-view KG with fine-grained instances and relations, and, (3) a set of cross-view links between two views of KG. $\text{Concept2Box}$ learns dual geometric embeddings where each concept is modeled as a box in the latent space and entities are learned as point vectors.}
	\label{fig:motivation}
\end{figure}

\section{Introduction} 
Knowledge graphs (KGs) like Freebase~\cite{freebase} and DBpedia~\cite{DBPedia} have motivated various knowledge-driven applications~\cite{qagnn,differentiable}. They contain structured and semantic information among nodes and relations where knowledge is instantiated as factual triples. One critical fact being ignored is that many KGs consist of two fundamentally different views as depicted in Figure~\ref{fig:motivation}: (1) an ontology-view component with high-level concepts and meta-relations, e.g., \textit{(``Singer'', ``is\_a'', ``Artist'')}, and (2) an instance-view component with specific entities and their relations, e.g., \textit{(``Taylor Swift'', ``has\_award'', ``Grammy Award'')}. Also, there are valuable cross-view links that connect ontological concepts with entities instantiated from them, which bridge the semantics of the two views.

 Modern KGs are usually far from complete as new knowledge is rapidly emerging~\cite{SS-AGA}. How to automatically harvest unknown knowledge from data has been an important research area for years. Knowledge graph embedding (KGE) methods achieve promising results for modeling KGs by learning low-dimensional vector representations of nodes and relations. However, most existing methods~\cite{transE,TransH,rotatE,hyperbolic_kg} have a major limitation of only considering KGs of a single view. The valuable semantics brought by jointly modeling the two views are omitted.
 In contrast, modeling two-view KGs in a holistic way has a key advantage: entity embeddings provide rich information for understanding corresponding ontological concepts; and, in turn, a concept embedding provides a high-level summary of linked entities, guiding the learning process of entities with only few relational facts.

Modeling two-view KG is non-trivial where the challenges are three-fold: First, there exists a significant structural difference between the two views, where the ontology-view KG usually forms a hierarchical structure with concept nodes of different granularity, and the instance-view KG usually follows a flat or cyclical structure~\cite{dgs} with each entity associated with a specific meaning. Second, concepts often exhibit complex relations such as intersect, contain and disjoint. Traditional vector-based embeddings~\cite{transE,rotatE} fail to adequately express them and capture interpretable probabilistic semantics to explain the granularity of concepts. Third, to bridge the knowledge from two views, effective semantic mappings from fine-grained entities to their corresponding concepts need to be carefully considered. For work that models two-view KGs, they either model concepts and entities all as points in the Euclidean space~\cite{junheng2019kdd} and in the product manifold space~\cite{m2gnn}, which can fail to capture the structural differences among the two views and result in suboptimal performance, or they embed concepts and entities separately as points in the hyperbolic and spherical space respectively~\cite{dgs}, which cannot provide interpretation towards concepts' granularity.  

To this end, we propose $\text{Concept2Box}$, a novel two-view KG embedding model for jointly learning the representations of concepts and instances using dual geometric objects.
Specifically, we propose to leverage the geometric objects of boxes (i.e. axis-aligned hyperrectangle) to embed concept nodes in the ontology-view KG.
This is motivated by the desirable geometric properties of boxes to represent more complex relationships among concepts such as intersect, contain, and disjoint, and also capture the hierarchy structure. For example, in Figure~\ref{fig:motivation} we can see \textit{City} is contained within \textit{Location}, denoting it's a sub-concept of \textit{Location}; \textit{Artist} and \textit{Scientist} intersects with each other, denoting they are correlated. In addition, box embeddings provide probabilistic measures of  concepts' granularity based on the volumes, such as \textit{Person} is a larger box than \textit{Scientist}. 
Entities in the instance-view KG are modeled with classic vector representations to capture their specific meanings. 
As we model concepts and entities using boxes and vectors respectively, we further design a new metric function to measure the distance from a vector to a box in order to bridge the semantics of the two views.
Based on this vector-to-box distance function, the proposed model is trained to jointly model the semantics of the instance-view KG, the ontology-view KG, and the cross-view links. Empirically, Concept2Box outperforms popular baselines on both public and our newly-created industrial recipe datasets.

Our contributions are as follows: (1) We propose a novel model for learning two-view KG representations by jointly embedding concepts and instances with different geometric objects. (2) We design a new metric function to measure the distance between concept boxes and entity vectors to bridge the two views. (3) We construct a new industrial-level recipe-related KG dataset. (4) Extensive experiments verify the effectiveness of Concept2Box.

\section{Related Work}
\subsection{Knowledge Graph Embedding (KGE)}
The majority of work on knowledge graph embeddings only considers a single view, where models learn the latent vector representations for nodes and relations and measure triple plausibility via varying score functions. Representatives include translation-based TransE~\cite{transE}, TransH~\cite{TransH}; rotation-based RotatE~\cite{rotatE}, and neural network-based ConvE~\cite{ConvE}. Most recently, some work employs graph neural networks (GNNs) to aggregate neighborhood information in KGs~\cite{SS-AGA,aaai_gnn}, which have shown superior performance on graph-structured
data by passing local message~\cite{LG-ODE,CG-ODE,hashtag,DyDiff-VAE}.
For work that pushes two-view or multiple KGs together~\cite{junheng2019kdd,SS-AGA}, they either model all nodes as points in the Euclidean space~\cite{junheng2019kdd,SS-AGA} and in the product manifold space~\cite{m2gnn}, or separately embed entities and concepts as points in two spaces respectively~\cite{dgs}. Our work uses boxes and vectors for embedding concepts and entities respectively, which not only gives better performance but also provides a probabilistic explanation of concepts' granularity based on box volumes.

\subsection{Geometric Representation Learning}
Representing elements with more complex geometric objects than vectors is an active area of study~\cite{xuelu_box}. Some representative models include Poincaré embeddings~\cite{nickel2017poincare}, which embeds data into the hyperbolic space with the inductive basis of negative curvature. Hyperbolic entailment cones~\cite{ganea2018hyperbolic} combine order embeddings with hyperbolic geometry, where relations are viewed as partial orders defined over nested geodesically convex cones. Elliptical distributions~\cite{elliiptical} model nodes as distributions and define distances based on the 2-Wasserstein metric. 
While they are promising for embedding hierarchical structures, they do not provide interpretable probabilistic semantics. Box embeddings not only demonstrate improved performance on modeling hierarchies but also model probabilities via box volumes to explain the granularity of concepts. 
\section{Preliminaries}
\subsection{Problem Formulation}~\label{sec:problem}
We consider a two-view knowledge graph $\mathcal{G}$ consisting of a set of entities $\mathcal{E}$, a set of concepts $\mathcal{C}$,  and a set of relations $\mathcal{R}=\mathcal{R}^{\mathcal{I}}\cup\mathcal{R}^{\mathcal{O}}\cup \mathcal{S}$, where $\mathcal{R}^{\mathcal{I}}, \mathcal{R}^{\mathcal{O}}$ are the entity-entity and concept-concept relation sets respectively. $\mathcal{S} = \{(e,c)\}$ is the cross-view entity-concept relation set with $e\in \mathcal{E}, c\in\mathcal{C}$. Entities and concepts are disjoint sets, i.e., $\mathcal{E}\cap\mathcal{C}=\emptyset$.

We denote the instance-view KG as $\mathcal{G}^{\mathcal{I}} := \{\mathcal{E},\mathcal{R}^{\mathcal{I}},\mathcal{T}^{\mathcal{I}}\}$ where $\mathcal{T}^{\mathcal{I}} =\{(h^{\mathcal{I}},r^{\mathcal{I}},t^{\mathcal{I}})\} $ are factual triples connecting head and tail entities $h^{\mathcal{I}}, t^{\mathcal{I}}\in \mathcal{E}$ with a relation $r^{\mathcal{I}}\in\mathcal{R}^{\mathcal{I}}$.
The instance-view triples can be any real-world instance associations and are often of a flat structure such as \textit{(``Taylor Swift'', ``has\_award'', ``Grammy Award'')}.
Similarly, the ontology-view KG is denoted as $\mathcal{G}^{\mathcal{O}} := \{\mathcal{C},\mathcal{R}^{\mathcal{O}},\mathcal{T}^{\mathcal{O}}\}$ where $\mathcal{T}^{\mathcal{O}} =\{(h^{\mathcal{O}},r^{\mathcal{O}},t^{\mathcal{O}})\} $ are triples connecting $h^{\mathcal{O}}, t^{\mathcal{O}}\in \mathcal{C}$ and $r^{\mathcal{O}}\in\mathcal{R}^{\mathcal{O}}$. 
The ontology-view triples describe the hierarchical structure among abstract concepts and their meta-relations such as \textit{(``Singer'', ``is\_a'', ``Artist'')}.

\textbf{Given:} a two-view KG $\mathcal{G}$, we want to \textbf{learn} a model which effectively embeds entities, concepts, and relations to latent representations with different geometric objects to better capture the structural differences of two views. We evaluate the quality of the learned embeddings on the KG completion task and concept linking task, described in Section~\ref{sec:exp}.

\subsection{Probabilistic Box Embedding}
Box embeddings represent an element as a hyperrectangle characterized with two parameters~\cite{box}: the minimum and the maximum corners $(\boldsymbol{x_m},\boldsymbol{x_M})\in\mathbb{R}^d$ where $x_m^i<x_M^i$ for each coordinates $i\in{1,2,\cdots d}$. Box volume is computed as the product of side-lengths $ \operatorname{Vol}(x)=\prod_{i=1}^d\left(x_M^i-x_m^i\right)$.
 Given a box space $\Omega_{\text {Box }} \subseteq \mathbb{R}^d$, the set of boxes $\mathcal{B}\left(\Omega_{\mathrm{Box}}\right)$ is closed under intersection~\cite{xuelu_box} and box volumes can be viewed as the unnormalized probabilities to interpret concepts' granularity. The intersection volume of two boxes is defined as  $\operatorname{Vol}(x \cap y)=\prod_i \max \left(\min \left(x_M^i, y_M^i\right)-\max \left(x_m^i, y_m^i\right), 0\right)$ and the conditional probability is computed as $ P(x|y) = \frac{\operatorname{Vol}(x \cap y)}{\operatorname{Vol}(y)},$ with value between 0 and 1~\cite{box}.

Directly computing the condition probabilities would pose difficulties during training. This can be caused by a variety of settings like when two boxes are disjoint but should overlap~\cite{xuelu_box}, there would be no gradient for some parameters. To alleviate this issue, we rely on Gumbel boxes~\cite{softbox}, where the min and max corners of boxes are modeled via Gumbel distributions:

\begin{equation}
    \begin{aligned}
\operatorname{Vol}(x) &=\prod_{i=1}^d\left[x_m^i, x_M^i\right] \quad \text { where } \\
x_m^i & \sim \operatorname{GumbelMax}\left(\mu_m^i, \beta\right), \\
x_M^i & \sim \operatorname { GumbelMin }\left(\mu_M^i, \beta\right),
\end{aligned}
\end{equation}
 $\mu_m^i,\mu_M^i$ are the minimum and maximum coordinates parameters; $\beta$ is the global variance. Gumbel distribution provides min/max stability~\cite{xuelu_box} and the expected volume is approximated as
\begin{equation}
\small
\begin{aligned}
&\mathbb{E}[\operatorname{Vol}(x)] \approx \prod_{i=1}^d \beta \log \left(1+\exp \left(\frac{\mu_M^i-\mu_m^i}{\beta}-2 \gamma\right)\right),
\end{aligned}
\end{equation}
where $\gamma$ is the Euler–Mascheroni constant~\cite{box4et}. The conditional probability becomes 
\begin{equation}
\mathbb{E}\left[P(x|y)\right] \approx \frac{\mathbb{E}[\operatorname{Vol}(x \cap y)]}{\mathbb{E}[\operatorname{Vol}(y)]}.
\end{equation}
This also leads to improved learning as the noise ensembles over a large collection of boxes which allows the model to escape plateaus in the loss function~\cite{xuelu,box4et}. We use this method when training box embeddings.

\section{Method}
\begin{figure*}[t]
    \centering
   \includegraphics[width=\linewidth]{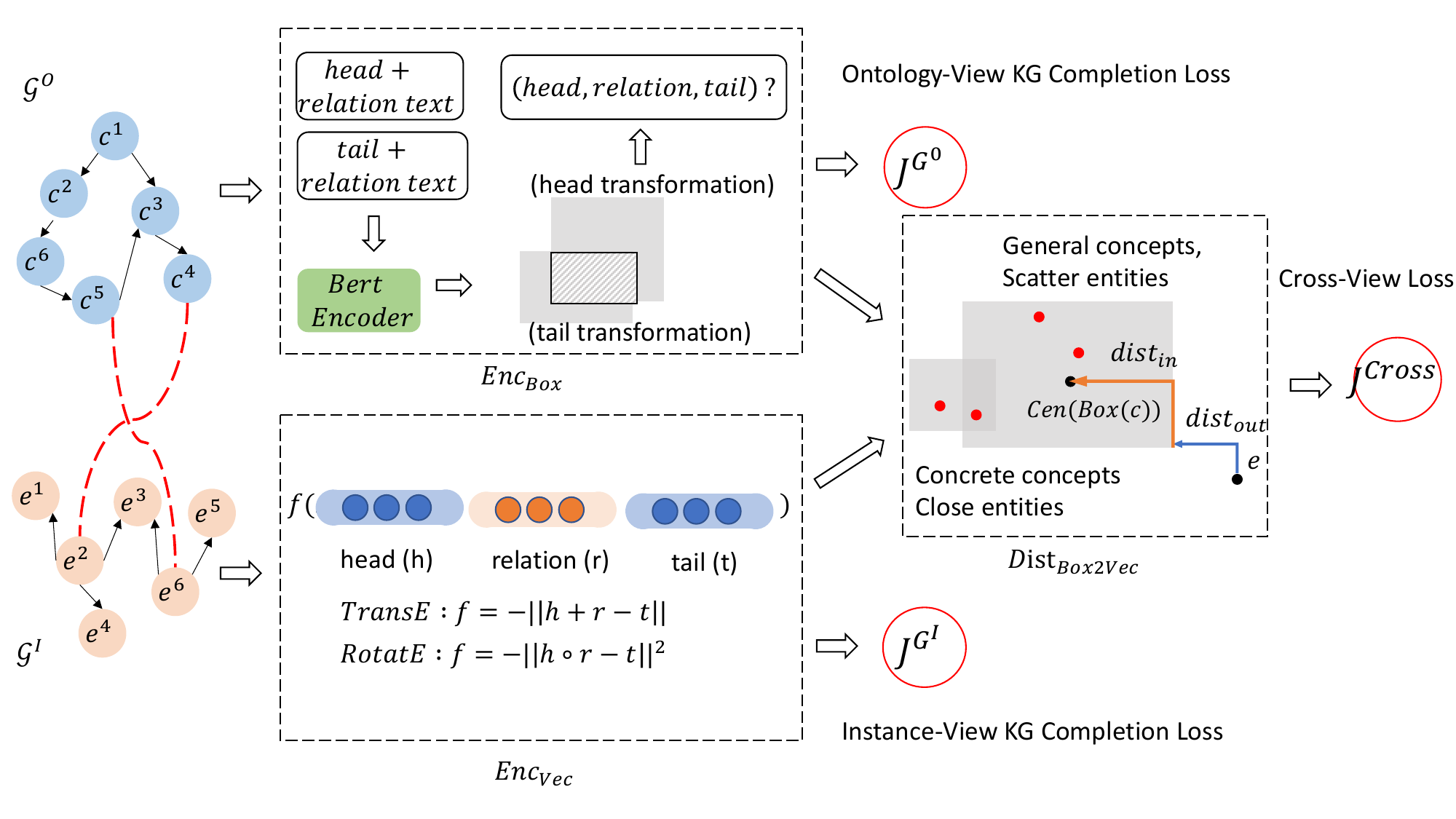}
  \caption{The overall framework of our proposed Concept2Box includes 3 modules. First, we employ probabilistic box embeddings to model the ontology-view KG, capturing the hierarchical structure and preserving the granularity of concepts (upper). Second, we model the instance-view KG by applying the vector-based KG embedding method (bottom). Third, to bridge these two views, we design a novel distance metric defined from a box to a vector (right). The model is learned by jointly optimizing three loss terms of each corresponding module.}
  \label{fig:framework}
\end{figure*}

In this section, we introduce Concept2Box, a novel KGE model to learn two-view knowledge graph embeddings with dual geometric representations. As shown in Figure~\ref{fig:framework}, Concept2Box consists of three main components: (1). The ontology-view box-based KG embedding module that captures the hierarchical structure and complex relations of concepts trained via the KG completion loss over $\mathcal{G}^\mathcal{O}$; (2). The instance-view vector-based KG embedding module trained via the KG completion loss over $\mathcal{G}^\mathcal{I}$. It has the flexibility to incorporate any off-the-shelf vector-based KGE methods~\cite{transE,rotatE}; (3). The cross-view module trained via the concept linking loss over $\mathcal{S}$. This module relies on a new distance metric for bridging the semantics between a vector and a box. We now present each component in detail.

\subsection{Ontology-View Box Embedding}
 For the ontology-view KG, we model each concept as a Gumbel box to capture the hierarchical structure and interpret concept granularity based on the box volume. Specifically, each Gumbel box x is parameterized using a center $\operatorname{cen}(x)\in\mathbb{R}^d$ and offset  $\operatorname{off}(x)\in\mathbb{R}^d_+$, and the minimum and maximum location parameters are given by  $\boldsymbol{\mu_{\mathrm{m}}}=\operatorname{cen}(x)-\operatorname{off}(x)$, $\boldsymbol{\mu_{\mathrm{M}}}=\operatorname{cen}(x)+\operatorname{off}(x)$.
To compute the triple score from box embeddings, we use the approximate conditional probability of observing a triple given its relation and tail, as shown 
\begin{equation}
\Scale[0.95]{
\phi(h^\mathcal{O},r^\mathcal{O},t^\mathcal{O})=\frac{\mathbb{E}\left[\operatorname{Vol}(f_{r^{\mathcal{O}}}(\text{Box}(h^\mathcal{O})) \cap f_{r^{\mathcal{O}}}(\text{Box}(t^\mathcal{O}))\right)]}{\mathbb{E}\left[\operatorname{Vol}(f_{r^{\mathcal{O}}}(\text{Box}(t^\mathcal{O})))\right]},}
\end{equation}
where $f_{r^{\mathcal{O}}}$ denotes the box transformation function to account for the influence of the relation $r^\mathcal{O}$. As shown in Figure~\ref{fig:motivation}, $f_\textit{lives\_in}(\textit{Artist})$, $f_\textit{is\_a}(\textit{Artist})$ are the transformed box embeddings of the concept \textit{Artist} in the context of the living locations and supertypes. The bounded score $\phi$ should be close to 1 for positive triples and 0 for negative ones.

Previous work~\citet{xuelu_box} defined the transformation function $f_r$ as shifting the original center and offset of the input box $x$: $\operatorname{cen}(f_r(x))=\operatorname{cen}(x)+\boldsymbol{\tau^r}$, $\operatorname{off}(f_r(x))=\operatorname{off}(x) \circ \boldsymbol{\Delta^r}$, where $\circ $ is the Hadamard product and $(\boldsymbol{\tau^r},\boldsymbol{\Delta^r})$ are relation-specific parameters. 
To further incorporate semantic information from the text of a concept $c$ and relation $r$, we propose a new transformation function as:
\begin{equation}
\small
\begin{aligned}
&\mathbf{h}^{[\text {CLS }]}=\operatorname{BERT}\left(c_{\text {text }}+\text { 'SEP' }+r_{\text {text }}\right)\\
&\operatorname{cen}\left(f_r(\operatorname{Box}(c))\right), \operatorname{off}\left(f_r(\operatorname{Box}(c))\right) = f_{\text{proj}}(\mathbf{h}^{[\text {CLS }]}).
\end{aligned}
\label{eq:box_encoder}
\end{equation}
Specifically, we first concatenate the textual content of the concept and relation for feeding into a pre-trained language model BERT~\cite{bert} that captures the rich semantics of the text.
We obtain the embedding by taking the hidden vector at the $[\text{CLS}]$ token. Then, we employ a projection function $f_{\text{proj}}$ to map the hidden vector $\mathbf{h}^{[\text {CLS }]} \in \mathbb{R}^{\ell}$ to the $\mathbb{R}^{2d}$ space, where $\ell$ is the hidden dimension of the BERT encoder, and $d$ is the dimension of the box space. In practice, we implement $f_{\text{proj}}$ as a simple two-layer Multi-Layer Perception (MLP) with a nonlinear activation function. We further split the hidden vector into two equal-length vectors as the center and offset of the transformed box.
\begin{remark}
Instead of encoding relation text first to get the shifting and scaling parameters and then conducting the box transformation as introduced in~\cite{xuelu}, we feed the concept and relation text jointly to the BERT encoder to allow interaction between them through the attention mechanism. The goal is to preserve both concept and relation semantics by understanding their textual information as a whole.
\end{remark}

We train the ontology-view KG embedding module with the binary cross entropy loss as below:

\begin{equation}
\Scale[1]{
\begin{aligned}
    \mathcal{J}^{\mathcal{G}^\mathcal{C}}=\sum_{(h^\mathcal{O}, r^\mathcal{O}, t^\mathcal{O}) \in \mathcal{T}^{\mathcal{O}}, \atop (\tilde{h}^\mathcal{O}, r^\mathcal{O}, \tilde{t}^\mathcal{O}) \notin \mathcal{T}^{\mathcal{O}}}|&\phi(h^\mathcal{O}, r^\mathcal{O}, t^\mathcal{O})-1|^2\\
    +\quad &|\phi\left(\tilde{h}^{\mathcal{O}}, r^\mathcal{O}, \tilde{t}^{\mathcal{O}}\right)|^2,
\end{aligned}
}
\label{eq:loss_O}
\end{equation}

\noindent where $(\tilde{h}^{\mathcal{O}}, r^\mathcal{O}, \tilde{t}^{\mathcal{O}})$ is a negative sampled triple obtained by replacing head or tail of the true triple $(h^O, r^O, t^O)$. We do not use the hinge loss defined  in vector-based KGE models~\cite{transE}, as the triple scores are computed from the conditional probability and strictly fall within 0 and 1.  

\subsection{Instance-View and Cross-View Modeling}\label{sec:cross}
Next, we introduce learning the instance-view KG and the cross-view module for bridging two views. 

\textbf{Instance-View KG Modeling.} As entities in the instance-view KG are fine-grained, we choose to model entities and relations as classic point vectors. We employ the standard hinge loss defined as follows to train the instance-view KGE module.

\begin{equation}
\Scale[0.9]{
\mathcal{J}^{\mathcal{G}^\mathcal{I}}=\sum_{\substack{\left(h^\mathcal{I}, r^\mathcal{I}, t^\mathcal{I}\right) \in \mathcal{T}^{\mathcal{I}} 
\\\left(\tilde{h}^{\mathcal{I}}, r^\mathcal{I}, \tilde{t}^{\mathcal{I}}\right) \notin \mathcal{T}^{\mathcal{I}}}}\left[
\begin{gathered}
f_{\text{vec}}\left(h^\mathcal{I}, r^\mathcal{I}, t^\mathcal{I}\right)\\
-f_{\text{vec}}\left(\tilde{h}^{\mathcal{I}}, r^\mathcal{I}, \tilde{t}^{\mathcal{I}}\right)\\
+\gamma^{\text{KG}}
\end{gathered}
\right]_{+},
}
\label{eq:loss_I}
\end{equation}

\noindent where $\gamma^{\text{KG}}>0$ is a positive margin, $f_{\text{vec}}$ is an arbitrary vector-based KGE model such as TransE~\cite{transE} and RotatE~\cite{rotatE}. And $(\tilde{h}^{\mathcal{I}}, r^\mathcal{I}, \tilde{t}^{\mathcal{I}})$ is a sampled negative triple.

\textbf{Cross-View KG Modeling.}
The goal of the cross-view module is to bridge the semantics between the entity embeddings and the concept embeddings by using the cross-view links. It forces any entity $e\in\mathcal{E}$ to be close
to its corresponding concept $c\in\mathcal{C}$. As we model concepts and entities with boxes and vectors respectively, existing distance measures between boxes (or between vectors) would be inappropriate for measuring the distance between them.
To tackle this challenge, we now define a new metric function to measure the distance from a vector to a box.
Our model is trained to minimize the distances for the concept-entity pairs in the cross-view links based on this metric function. Specifically, given a concept $c$ and an entity $e$, if we denote the minimum and maximum location parameters of the concept box as $\boldsymbol{\mu_m},\boldsymbol{\mu_M}$, and the vector for the entity as $\boldsymbol{e}$, 
we define the distance function $f_d$ as:
\begin{equation}
\Scale[0.825]{
\begin{aligned}
&f_d(e, c)=\operatorname{dist}_{\text {out}}(e, c)+\alpha \cdot \operatorname{dist}_{\text {in}}(e,c)\\
&\operatorname{dist}_{\text {out}}(e, c)=\left\|\operatorname{Max}\left(\boldsymbol{e}-\boldsymbol{\mu_M}, \mathbf{0}\right)+\operatorname{Max}\left(\boldsymbol{\mu_m}-\boldsymbol{e}, \mathbf{0}\right)\right\|_1\\
&\operatorname{dist}_{\text {in}}(e,c)=\left\|\operatorname{Cen}(\operatorname{Box}(c))-\operatorname{Min}\left(\boldsymbol{\mu_M}, \operatorname{Max}\left(\boldsymbol{\mu_m},\boldsymbol{e}\right)\right)\right\|_1
\end{aligned}
}
\end{equation}

\noindent where $\operatorname{dist}_{\text {out }}$ corresponds to the distance between the entity and closest corner of the box; $\operatorname{dist}_{\text {in}}$ corresponds to the distance between the center of the
box and its side or the entity itself if the entity is inside the box~\cite{query2box}. $0 <\alpha< 1$ is a fixed scalar to balance the outside and inside distances, with the goal to punish more if the entity falls outside of the concept, and less if it is within the box but not close to the box center. An example is shown in Figure~\ref{fig:framework}. The proposed vector-to-box distance metric is nonnegative and commutative.

However, there is one drawback to the above formulation: different boxes should have customized volumes reflecting their granularity. A larger box may be a more general concept (e.g., \textit{Person}),
so it is suboptimal for all paired entities to be as close as to its center, in contrast to more concrete concepts like \textit{Singer}, as illustrated in Figure~\ref{fig:framework}. Moreover, for instances that belong to multiple concepts, roughly pushing them to be equally close to the centers of relevant concepts can result in poor embedding results. Consider we have two recipe concepts \textit{KungPao chicken} and \textit{Curry chicken}, suppose \textit{KungPao Chicken} is more complicated and requires more ingredients. If we roughly push cooking wine, salt, and chicken breast these shared ingredient instances to be as close as to both of their centers, it will possibly result in forcing the centers and volume of the two recipes to be very similar to each other.
Therefore, we propose to associate the balancing scalar $\alpha$ to be related to each concept box volume. Specifically, we define $\alpha(x)$ with regard to a box $x$ as follows:

\begin{equation}
    \begin{aligned}
    \alpha(x) = \alpha \cdot \text{Sigmoid}(\frac{1}{\text{Vol}(x)}).
    \end{aligned}
    \label{eq:balancing}
\end{equation}
We multiply the original constant coefficient $\alpha$ with a box-specific value negatively proportional to its volume. The idea is to force entities to be closer to the box center when the box has a smaller volume.

The cross-view module is trained by minimizing the following loss with negative sampling:
\begin{equation}
\small
\mathcal{J}^{\text {Cross }}=\sum_{(e, c) \in \mathcal{S} \atop (e\prime,c\prime) \notin \mathcal{S}}
\left[\sigma(f_d(e, c))-\sigma\left(f_d\left(e^{\prime}, c^{\prime}\right)\right)+\gamma^{\text{Cross}}\right],
\label{eq:loss_cross}
\end{equation}
\noindent where $\gamma^{\text{Cross}}$ represents a fixed scalar margin.

\subsection{Overall Training Objective}
Now that we have presented all model components, the overall loss function is a linear combination of the instance-view and ontology-view KG completion loss, and the cross-view loss, as shown below:
\begin{equation}
\mathcal{J}=\mathcal{J}^{\mathcal{G}^{\mathcal{O}}} + \lambda_1 \mathcal{J}^{\mathcal{G}^{\mathcal {I}} }+ \lambda_2 \mathcal{J}^{\text {Cross }},
    \label{eq:loss_all}
\end{equation}
where $\lambda_1, \lambda_2 > 0$ are two positive hyperparameters to balance the scale of the three loss terms. 

\subsection{Time Complexity Analysis}
The time complexity of Concept2Box is $\mathcal{O}(3d|\mathcal{T}^\mathcal{O}| + d|\mathcal{T}^\mathcal{I}| + 5d|\mathcal{S}|)$, where $\mathcal{T}^\mathcal{O}, \mathcal{T}^\mathcal{I}, \mathcal{S}$  are the ontology-view KG triples, instance-view KG triples, and cross-view set defined in Sec~\ref{sec:problem}. Here $d$ is the embedding dimension of vectors. The time complexity of other vector-based methods is $\mathcal{O}(d|\mathcal{T}^\mathcal{O}| + d|\mathcal{T}^\mathcal{I}| + d|\mathcal{S}|)$ which is asymptotically the same as Concept2Box.

\begin{table*}[htb]
\centering
\Scale[0.825]{
\begin{tabular}{c|ccc|ccc|c}
 \toprule
\hline
Dataset & \multicolumn{3}{c|}{Instance-View KG}   & \multicolumn{3}{c|}{Ontology-View KG}  & Cross-View Links \\ \hline
        & \# Entities & \# Relations & \# Triples & \# Concepts & \# Relations & \# Triples & \# Links         \\ \hline
DBpedia & 111,762     & 305          & 863,643    & 174         & 20          & 763        & 99,748           \\ \hline
Recipe  & 21,457      & 9            & 200,288    & 32,922      & 5           & 445,632    & 11,474           \\ \hline
\bottomrule
\end{tabular}
}
\caption{Statistics of the DBpedia and Recipe datasets.}
\label{table:data_stats}
\end{table*}

\section{Experiments}\label{sec:exp}
In this section, we introduce our empirical results against existing baselines to evaluate our model performance. We start with the datasets and baselines introduction and then conduct the performance evaluation. Finally, we conduct a model generalization experiment and a case study to better illustrate our model design.
\subsection{Dataset and Baselines}
We conducted experiments over two datasets: one public dataset from DBpedia~\cite{junheng2019kdd}, which describes general concepts and fine-grained entities crawled from DBpedia. We also created a new recipe-related dataset, where concepts are recipes, general ingredient and utensil names, etc, and the entities are specific products for cooking each recipe searched via Amazon.com, along with some selected attributes such as brand. The detailed data structure can be found in Appendix~\ref{appendix1}. The statistics of the two datasets are shown in Table~\ref{table:data_stats}.

We compare against both single-view KGE models and two-view KGE models as follows:

\noindent $\bullet$\textbf{Vector-based KGE models:}: Single-view vector-based KGE models that represent triples as vectors, including TransE~\cite{transE}, RotatE~\cite{rotatE},DistMult~\cite{dismult}, and ComplexE~\cite{complexE}. 
    
\noindent $\bullet$ \textbf{KG$\-$Bert}~\cite{KGbert}: Single-view KG embedding method which employs BERT to incorporate 
text information of triples. 
    
\noindent $\bullet$ \textbf{Box4ET}~\cite{box4et}: Single-view KGE model that represents triples using boxes.  

\noindent $\bullet$\textbf{Hyperbolic GCN}~\cite{hyperbolic_gcn}: Single-view KGE method that represents triples in the hyperbolic space.

\noindent $\bullet$\textbf{JOIE}~\cite{junheng2019kdd}: Two-view KGE method that embeds entities and concepts all as vectors.

\subsection{KG Completion Performance}

\begin{table*}[htb]
\resizebox{1.0\linewidth}{!}{
\begin{tabular}{c|cccccc|cccccc}
 \toprule
\hline
                                 & \multicolumn{6}{c|}{Recipe}                                                                                                                                                                                           & \multicolumn{6}{c}{DBpedia}                                                                                              \\ \cline{2-13} 
                                 & \multicolumn{3}{c|}{$\mathcal{G}^{\mathcal{I}}$ KG Completion}                                                                         & \multicolumn{3}{c|}{$\mathcal{G}^{\mathcal{O}}$ KG Completion}                                           & \multicolumn{3}{c|}{$\mathcal{G}^{\mathcal{I}}$ KG Completion}                      & \multicolumn{3}{c}{$\mathcal{G}^{\mathcal{O}}$KG Completion}  \\ \hline
                                 & MRR                          & H@1                          & \multicolumn{1}{c|}{H@10}                                  & MRR                          & H@1                          & H@10                         & MRR            & H@1            & \multicolumn{1}{c|}{H@10}           & MRR            & H@1            & H@10           \\
TransE                           & {\color[HTML]{343434} 0.28} & {\color[HTML]{343434} 26.31} & \multicolumn{1}{c|}{{\color[HTML]{343434} 38.59}} & {\color[HTML]{343434} 0.18} & {\color[HTML]{343434} 11.24} & {\color[HTML]{343434} 30.67} & 0.32          & 22.70          & \multicolumn{1}{c|}{48.12}          & 0.54          & 47.90          & 61.84          \\
RotatE                           & 0.30                        & 28.31                        & \multicolumn{1}{c|}{40.01}                                 & 0.22                        & 15.68                        & 35.47                        & 0.36          & 29.31          & \multicolumn{1}{c|}{54.60}          & 0.56          & 49.16          & 68.19          \\
DistMult                         & 0.29                        & 27.33                        & \multicolumn{1}{c|}{39.04}                                 & 0.18                        & 12.37                        & 32.04                        & 0.28          & 27.24          & \multicolumn{1}{c|}{29.70}          & 0.50          & 45.52          & 64.73          \\
ComplexE                         & 0.27                        & 26.58                        & \multicolumn{1}{c|}{37.99}                                 & 0.19                        & 14.76                        & 34.02                        & 0.32          & 27.39          & \multicolumn{1}{c|}{46.63}          & 0.55          & 47.80          & 62.23          \\
JOIE                             & 0.41                        & 29.73                        & \multicolumn{1}{c|}{60.14}                                 & 0.36                        & 22.58                        & 53.62                        & 0.48          & 35.21          & \multicolumn{1}{c|}{72.38}          & 0.60          & 52.48          & 79.71          \\
Hyperbolic GCN                   & 0.36                        & 28.58                        & \multicolumn{1}{c|}{52.48}                                 & 0.25                        & 16.04                        & 40.35                        & 0.30          & 33.68               & \multicolumn{1}{c|}{46.72}          & 0.54          & 47.59          & 62.11          \\
\multicolumn{1}{c|}{KG-Bert}     & 0.30                        & 29.92                        & \multicolumn{1}{c|}{51.32}                                 & 0.24                        & 19.04                        & 38.78                        & 0.39          & 31.10          & \multicolumn{1}{c|}{60.46}          & 0.63          & 55.69          & 81.07          \\
\multicolumn{1}{c|}{Box4ET}      & 0.42                        & 28.81                        & \multicolumn{1}{c|}{59.88}                                 & 0.35                        & 23.01                        & 53.79                        & 0.42          & 33.69          & \multicolumn{1}{c|}{68.12}          & 0.64          & 55.89          & 81.45          \\
\multicolumn{1}{c|}{Concept2Box} & \textbf{0.44}               & \textbf{30.33}               & \multicolumn{1}{c|}{\textbf{62.01}}                        & \textbf{0.37}               & \textbf{23.16}               & \textbf{54.92}               & \textbf{0.51} & \textbf{36.52} & \multicolumn{1}{c|}{\textbf{73.11}} & \textbf{0.65} & \textbf{56.82} & \textbf{83.01} \\ \hline
\bottomrule
\end{tabular}}
\caption{Results of KG completion task. Best results are bold-faced}
\label{table:kg_results}
\end{table*}

The KG completion (KGC) task is to predict missing triples based on the learned embeddings, which helps to facilitate the knowledge learning process in KG construction. Without loss of generality, we discuss the case of predicting missing tails,
which we also refer to as a query $q = (h, r, ?t)$~\cite{SS-AGA,xuelu_box}. We conduct the KGC task for both views against baselines.

\textbf{Evaluation Protocol.} We evaluate the KGC performance using mean reciprocal ranks (MRR), accuracy (Hits@1) and the proportion of correct answers ranked within the top 10 (Hits@10). All three metrics are preferred to be higher, so as to indicate better KGC performance. For both $\mathcal{G}^{\mathcal{I}}$ and $\mathcal{G}^{\mathcal{O}}$, we randomly split the triples into three parts: 80$\%$ for training, $10\%$ for validation and $10\%$  for testing. During model training and testing, we utilize all cross-view links $\mathcal{S}$.

\textbf{Results.} The evaluation results are shown in Table~\ref{table:kg_results}. Firstly, by comparing the average performance between single-view and two-view KG models, we can observe that two-view methods are able to achieve better results for the KGC task. This indicates that the intuition behind modeling two-view KGs to conduct KG completion is indeed beneficial. Secondly, by comparing vector-based methods with other geometric representation-based methods such as Hyperbolic GCN, there is a performance improvement on most metrics, indicating that simple vector representations have its limitation, especially in capturing the hierarchical structure among concepts (shown by the larger performance gap in $\mathcal{G}^{\mathcal{O}}$). Our Concept2Box is able to achieve higher performance in most cases, verifying the effectiveness of our design.

\subsection{Concept Linking Performance}\label{sec:concept_linking}
The concept linking task predicts the associating concepts of given entities, where each entity can be potentially mapped to multiple concepts. It tests the quality of the learned embeddings.

\textbf{Evaluation Protocol.} We split the cross-view links $\mathcal{S}$ into $80\%$, $10\%$, $10\%$ for training, validation, testing respectively. We utilize all the triples in both views for training and testing. For each concept-entity pair, we rank all concept candidates for the given entity based on the distance metric from a box to a vector introduced in Sec~\ref{sec:cross}, and report MRR, Hits@1 and Hits@3 for evaluation.

\textbf{Results.}
The evaluation results are shown in Table~\ref{table:concept_results}. We can see that Concept2Box is able to achieve the highest performance across most metrics, indicating its effectiveness. By comparing with Box4ET where both entities and concepts are modeled as boxes, our Concept2Box performs better, which indicates our intuition that entities and concepts are indeed two fundamentally different types of nodes and should be modeled with different geometric objects. Finally, we observe that Box4ET, and Concept2Box are able to surpass methods that are not using box embeddings, which shows the advantage of box embeddings in learning the hierarchical structure and complex behaviors of concepts with varying granularity.

\begin{table}[htb] 
\Scale[0.8]{
\fontsize{10}{10}\selectfont
\begin{tabular}{c|ccc|ccc}
\toprule
\hline
Datasets       & \multicolumn{3}{c|}{Recipe}                      & \multicolumn{3}{c}{DBPedia}                      \\\cline{2-7}
Metrics        & MRR            & Acc.           & Hit@3          & MRR            & Acc.           & Hit@3          \\ \hline
TransE         & 0.23          & 21.36          & 30.68          & 0.53          & 43.67          & 60.78          \\
RotatE         & 0.25          & 23.01          & 33.34          & 0.72          & 61.48          & 75.67          \\
DistMult       & 0.24          & 22.34          & 31.05          & 0.55          & 49.83          & 68.01          \\
ComplexE       & 0.22          & 22.50          & 31.45          & 0.58          & 55.07          & 70.17          \\
JOIE           & 0.57          & 54.58          & 66.39          & 0.85          & 75.58          & 96.02          \\
Hyper. GCN & 0.60          & 53.49          & 67.03          & 0.86          & 76.11          & 96.50          \\
Box4ET         & 0.59          & 54.36          & 68.88          & \textbf{0.88} & 77.04          & 97.38          \\
Concept2Box    & \textbf{0.61} & \textbf{55.33} & \textbf{69.01} & 0.87          & \textbf{78.09} & \textbf{97.44} \\ \hline
\bottomrule
\end{tabular}
}
\caption{Results of concept linking task.}
\label{table:concept_results}
\end{table}

\textbf{Abaltion Studies.} To evaluate the effectiveness of our model design,
we conduct an ablation study on the DBpedia dataset as shown in Table~\ref{table:ablation}. First, to show the effect of joint modeling the text semantics of relations and concepts for generating the relation-transformed box embeddings as in Eqn~\ref{eq:box_encoder}, we compare with separately modeling relations as box shifting and scaling introduced in~\cite{xuelu_box}. The latter one gives poorer results, indicating the effectiveness of our proposed box encoder. Next, we study the effect of the volume-based balancing scalar defined in Eqn~\ref{eq:balancing}. Compared with a fixed balancing scalar for all boxes, Concept2Box is able to generate better results.

\begin{table}[htb] 
\Scale[0.55]{
\begin{tabular}{l|ccc|ccc|ccc}
\toprule
\hline
\multicolumn{1}{c|}{}                                                   & \multicolumn{3}{c|}{$\mathcal{G}^{\mathcal{I}}$ KG Completion} & \multicolumn{3}{c|}{$\mathcal{G}^{\mathcal{O}}$ KG Completion} & \multicolumn{3}{c}{Concept Linking}              \\
\multicolumn{1}{c|}{}                                                   & MRR            & H@1            & H@10           & MRR            & H@1            & H@10           & MRR            & H@1            & H@3            \\ \hline
Concept2Box                                                             & \textbf{0.506} & \textbf{36.52} & \textbf{73.11} & \textbf{0.644} & \textbf{56.82} & \textbf{83.01} & \textbf{0.874} & \textbf{78.09} & \textbf{97.44} \\ \cline{1-1}
\begin{tabular}[c]{@{}l@{}}- relation \\  shift\&scale\end{tabular}     & 0.483          & 34.99          & 72.18          & 0.631          & 55.34          & 81.97          & 0.865          & 77.64          & 96.34          \\ \cline{1-1}
\begin{tabular}[c]{@{}l@{}}- fixed distance \\ coefficient\end{tabular} & 0.479          & 35.13          & 71.88          & 0.635          & 54.98          & 82.33          & 0.854          & 76.91          & 96.13          \\ \hline \bottomrule
\end{tabular}
}
\caption{Results of ablation study on DBpedia.}
\label{table:ablation}
\end{table}

\begin{figure*}[htbp]
    \centering
\includegraphics[width=\linewidth]{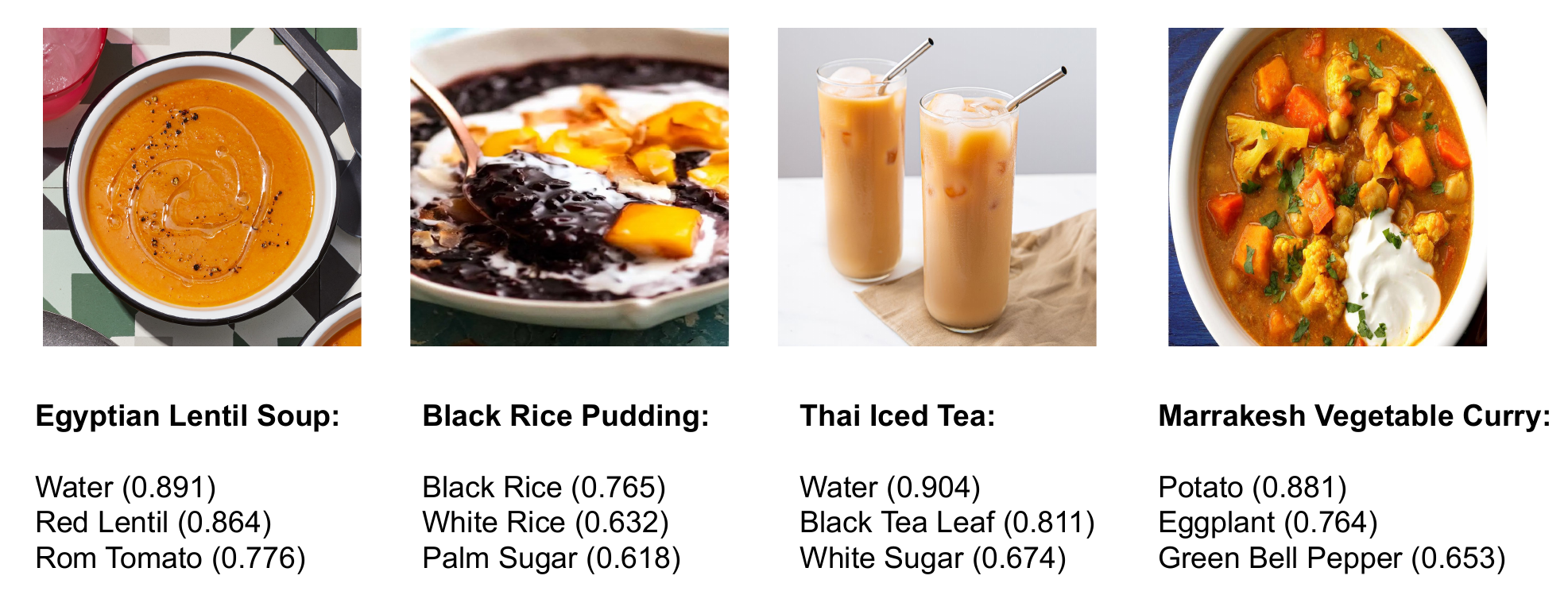}
  \caption{Case Study on the Recipe Dataset.}
  \label{fig:case study}
  \vspace{-10pt}
\end{figure*}
\subsection{Model Generalization}
We test our model's generalization ability by training our model on the Recipe dataset and evaluating it on the concept linking task with  recipe-product pairs collected from other online resources. 
Specifically, in the evaluation data, all products are known entities in the Recipe dataset during training, where we can directly obtain their embeddings. For recipes, only 3$\%$ of them are observed, and for unseen recipes, we generate their box embeddings by sending their title as input to the box encoder. We divide these recipe-product pairs into two groups based on whether the recipes are known.
We use the first group with known recipes to test the generalization ability in the transductive setting: When test data follows a different distribution as opposed to the training data, how does the model perform? It differs from Sec~\ref{sec:concept_linking} in that the test data share the same distribution with the training data as we randomly split them from the same dataset. We use the second group with unknown recipes to test in the inductive setting: When unknown concepts are sent as input, how does the model perform?

\begin{table}[]
\scalebox{0.72}{
\begin{tabular}{l|lll|lll}
\toprule
\hline
                                                              & \multicolumn{3}{l|}{Known Recipes} & \multicolumn{3}{l}{Unknown Recipes} \\ \hline
\begin{tabular}[c]{@{}l@{}}\# types *\\ \# items\end{tabular} & 10*12      & 12*10     & 24*5      & 10*12      & 12*10      & 24*5      \\ \hline
JOIE                                                          & 68.33      & 69.17     & 68.45     & 57.34      & 57.56      & 56.38     \\
Concept2Box                                                   & \textbf{71.97}      & \textbf{73.33}     & \textbf{69.43}     & \textbf{61.32}      & \textbf{66.77}      & \textbf{59.68 }    \\ \hline \bottomrule
\end{tabular}}
\caption{H@120 with different number of item types.}
\label{table:diversity}
\end{table}

\textbf{Diversity-Aware Evaluation.} To verify our learned embeddings can capture complex relations and hierarchy structure among concepts, we use diversity-aware evaluation for the generalization experiment similar to that in~\cite{junhengCIKM}. The idea is that for each recipe, a more preferred ranking list is one consisting of products from multiple necessary ingredients, instead of the one with products all from one ingredient. Since we model recipes and ingredients as concepts, and products as entities, for each recipe, we first extract its top $K$ ingredients (types) from $\mathcal{G}^{\mathcal{O}}$, and then for each of the $K$ ingredient we extract the top $M$ products (items). We set $M\times K = 120$ and let $K = 10, 12, 24$.

\textbf{Results.} Table~\ref{table:diversity} shows the results of H@120 among two-view KGE methods. We can see that Concept2Box is able to achieve the best across different settings, showing its great generalization ability. Note that when changing the number of types, Concept2Box is able to bring more performance gain than JOIE. This can be understood as the hierarchy structure of concepts is well captured by the box embeddings, when properly selecting the number of types, we first narrow down the concept (recipe) to the set of related concepts (ingredients) for a better understanding, which will generate better results.

\subsection{Case Study}

We conduct a case study to investigate the semantic meaning of learned box embeddings of concepts. We select recipe concepts and show the top 3 concepts that have the largest intersection with them after the relation transformation of "using ingredients". As shown in Figure~\ref{fig:case study}, the dominant ingredients for each  recipe are at the top of the list, verifying the effectiveness of Concept2Box.  Also, we found that \textit{Egyptian lentil Soup} has a larger intersection with \textit{Black Rice Pudding} than with \textit{Thai Iced Tea}, since the first two are dishes with more similarities and the third one is beverage. We also examine the box volume of concepts which reflects their granularity. Among ingredient concepts, we observe that Water, Sugar, and Rice are of the largest volumes, indicating they are common ingredients shared across recipes. This observation confirms that the box
volume effectively represents probabilistic semantics,
which we believe to be a reason for the performance
improvement.

\section{Conclusion}
In this paper, we propose {\it Concept2Box}, a novel two-view knowledge graph embedding method with dual geometric representations. We model high-level concepts as boxes to capture the hierarchical structure and complex relations among them and reflect concept granularity based on box volumes. For the instance-view KG, we model fine-grained entities as vectors and propose a novel metric function to define the distance between a box to a vector to bridge the semantics of the two views. Concept2Box is jointly trained to model the instance-view KG, the ontology-view KG, and the cross-view links. Empirical experiment results on two real-world datasets including a newly-created recipe dataset verify the effectiveness of Concept2Box.

\section{Limitations}   
Our model is developed to tackle the structural difference between the ontology and instance views of a KG. However, many modern KGs are multilingual, where different portions of a KG may not only differ in structure but also differ in the text semantics. How to jointly capture these differences remain unsolved. Also since box embeddings naturally provide interpretability to the granularity of the learned concepts, how to use the current model to discover unknown concepts from these embeddings is also challenging.

\section{Ethical Impact}      
Our paper proposed Concept2Box, a novel two-view knowledge graph embedding model with dual geometric representations. Concept2Box neither introduces any social/ethical bias to the
model nor amplifies any bias in the data. For the created recipe dataset, we did not include any customers'/sellers' personal information and only preserved information related to products' attributes and their relations. Our model implementation is built upon public libraries in Pytorch. Therefore, was e do not foresee any direct social consequences or ethical issues.

\section{Acknowledgement}
This work was partially supported by NSF 1829071, 2106859, 2200274, 2211557, 1937599, 2119643, 2303037, DARPA \#HR00112290103/HR0011260656, NASA, Amazon Research Awards, and an NEC research award.

\bibliographystyle{acl_natbib}
\bibliography{custom}

\begin{thebibliography}{34}
\expandafter\ifx\csname natexlab\endcsname\relax\def\natexlab#1{#1}\fi

\bibitem[{Bollacker et~al.(2008)Bollacker, Evans, Paritosh, Sturge, and
  Taylor}]{freebase}
Kurt Bollacker, Colin Evans, Praveen Paritosh, Tim Sturge, and Jamie Taylor.
  2008.
\newblock Freebase: A collaboratively created graph database for structuring
  human knowledge.
\newblock In \emph{Proceedings of the 2008 ACM SIGMOD International Conference
  on Management of Data}, SIGMOD '08, page 1247–1250.

\bibitem[{Bordes et~al.(2013)Bordes, Usunier, Garcia-Dur\'{a}n, Weston, and
  Yakhnenko}]{transE}
Antoine Bordes, Nicolas Usunier, Alberto Garcia-Dur\'{a}n, Jason Weston, and
  Oksana Yakhnenko. 2013.
\newblock Translating embeddings for modeling multi-relational data.
\newblock In \emph{Proceedings of the 26th International Conference on Neural
  Information Processing Systems - Volume 2}, page 2787–2795.

\bibitem[{Chami et~al.(2019)Chami, Ying, R{\'e}, and Leskovec}]{hyperbolic_gcn}
Ines Chami, Zhitao Ying, Christopher R{\'e}, and Jure Leskovec. 2019.
\newblock Hyperbolic graph convolutional neural networks.
\newblock In \emph{Advances in Neural Information Processing Systems}, pages
  4869--4880.

\bibitem[{Chen et~al.(2021)Chen, Boratko, Chen, Dasgupta, Li, and
  McCallum}]{xuelu_box}
Xuelu Chen, Michael Boratko, Muhao Chen, Shib~Sankar Dasgupta, Xiang~Lorraine
  Li, and Andrew McCallum. 2021.
\newblock Probabilistic box embeddings for uncertain knowledge graph reasoning.
\newblock In \emph{Proceedings of the 19th Annual Conference of the North
  American Chapter of the Association for Computational Linguistics (NAACL)}.

\bibitem[{Chen et~al.(2020)Chen, Chen, Fan, Uppunda, Sun, and Zaniolo}]{xuelu}
Xuelu Chen, Muhao Chen, Changjun Fan, Ankith Uppunda, Yizhou Sun, and Carlo
  Zaniolo. 2020.
\newblock Multilingual knowledge graph completion via ensemble knowledge
  transfer.
\newblock In \emph{Proceedings of the 2020 Conference on Empirical Methods in
  Natural Language Processing: Findings}, pages 3227--3238.

\bibitem[{Dasgupta et~al.(2020)Dasgupta, Boratko, Zhang, Vilnis, Li, and
  McCallum}]{softbox}
Shib Dasgupta, Michael Boratko, Dongxu Zhang, Luke Vilnis, Xiang Li, and Andrew
  McCallum. 2020.
\newblock Improving local identifiability in probabilistic box embeddings.
\newblock \emph{Advances in Neural Information Processing Systems},
  33:182--192.

\bibitem[{Dettmers et~al.(2018)Dettmers, Pasquale, Pontus, and Riedel}]{ConvE}
Tim Dettmers, Minervini Pasquale, Stenetorp Pontus, and Sebastian Riedel. 2018.
\newblock Convolutional 2d knowledge graph embeddings.
\newblock In \emph{Proceedings of the 32th AAAI Conference on Artificial
  Intelligence}, pages 1811--1818.

\bibitem[{Devlin et~al.(2019)Devlin, Chang, Lee, and Toutanova}]{bert}
Jacob Devlin, Ming-Wei Chang, Kenton Lee, and Kristina Toutanova. 2019.
\newblock {BERT}: Pre-training of deep bidirectional transformers for language
  understanding.
\newblock In \emph{Proceedings of the 2019 Conference of the North {A}merican
  Chapter of the Association for Computational Linguistics: Human Language
  Technologies, Volume 1 (Long and Short Papers)}, pages 4171--4186.

\bibitem[{Ganea et~al.(2018)Ganea, B{\'e}cigneul, and
  Hofmann}]{ganea2018hyperbolic}
Octavian Ganea, Gary B{\'e}cigneul, and Thomas Hofmann. 2018.
\newblock Hyperbolic entailment cones for learning hierarchical embeddings.
\newblock In \emph{International Conference on Machine Learning}, pages
  1646--1655. PMLR.

\bibitem[{Hao et~al.(2019)Hao, Chen, Yu, Sun, and Wang}]{junheng2019kdd}
Junheng Hao, Muhao Chen, Wenchao Yu, Yizhou Sun, and Wei Wang. 2019.
\newblock Universal representation learning of knowledge bases by jointly
  embedding instances and ontological concepts.
\newblock In \emph{Proceedings of the 25th ACM SIGKDD International Conference
  on Knowledge Discovery and Data Mining}, KDD '19, page 1709–1719.

\bibitem[{Hao et~al.(2020)Hao, Zhao, Li, Dong, Faloutsos, Sun, and
  Wang}]{junhengCIKM}
Junheng Hao, Tong Zhao, Jin Li, Xin~Luna Dong, Christos Faloutsos, Yizhou Sun,
  and Wei Wang. 2020.
\newblock P-companion: A principled framework for diversified complementary
  product recommendation.
\newblock In \emph{Proceedings of the 29th ACM International Conference on
  Information Knowledge Management}, CIKM '20, page 2517–2524.

\bibitem[{Huang et~al.(2022)Huang, Li, Jiang, Cao, Lu, Yin, Subbian, Sun, and
  Wang}]{SS-AGA}
Zijie Huang, Zheng Li, Haoming Jiang, Tianyu Cao, Hanqing Lu, Bing Yin, Karthik
  Subbian, Yizhou Sun, and Wei Wang. 2022.
\newblock Multilingual knowledge graph completion with self-supervised adaptive
  graph alignment.
\newblock In \emph{Annual Meeting of the Association for Computational
  Linguistics (ACL)}.

\bibitem[{Huang et~al.(2020)Huang, Sun, and Wang}]{LG-ODE}
Zijie Huang, Yizhou Sun, and Wei Wang. 2020.
\newblock Learning continuous system dynamics from irregularly-sampled partial
  observations.
\newblock In \emph{Advances in Neural Information Processing Systems}.

\bibitem[{Huang et~al.(2021)Huang, Sun, and Wang}]{CG-ODE}
Zijie Huang, Yizhou Sun, and Wei Wang. 2021.
\newblock Coupled graph ode for learning interacting system dynamics.
\newblock In \emph{Proceedings of the 27th ACM SIGKDD Conference on Knowledge
  Discovery and Data Mining}.

\bibitem[{Iyer et~al.(2022)Iyer, Bai, Wang, and Sun}]{dgs}
Roshni~G. Iyer, Yunsheng Bai, Wei Wang, and Yizhou Sun. 2022.
\newblock Dual-geometric space embedding model for two-view knowledge graphs.
\newblock In \emph{Proceedings of the 28th ACM SIGKDD Conference on Knowledge
  Discovery and Data Mining}, KDD '22, page 676–686.

\bibitem[{Javari et~al.(2020)Javari, He, Huang, Jeetu, and
  Chen-Chuan~Chang}]{hashtag}
Amin Javari, Zhankui He, Zijie Huang, Raj Jeetu, and Kevin Chen-Chuan~Chang.
  2020.
\newblock Weakly supervised attention for hashtag recommendation using graph
  data.
\newblock In \emph{Proceedings of The Web Conference 2020}, WWW '20.

\bibitem[{Kingma and Ba(2014)}]{adam}
Diederik~P Kingma and Jimmy Ba. 2014.
\newblock Adam: A method for stochastic optimization.
\newblock In \emph{International Conference on Learning Representations}.

\bibitem[{Lehmann et~al.(2015)Lehmann, Isele, Jakob, Jentzsch, Kontokostas,
  Mendes, Hellmann, Morsey, Kleef, and Auer}]{DBPedia}
Jens Lehmann, Robert Isele, Max Jakob, Anja Jentzsch, Dimitris Kontokostas,
  Pablo~N Mendes, Sebastian Hellmann, Mohamed Morsey, Patrick~Van Kleef, and
  Soren Auer. 2015.
\newblock Dbpedia–a large-scale, multilingual knowledge base extracted from
  wikipedia.
\newblock In \emph{Semantic Web}, pages 167--195.

\bibitem[{Lin et~al.(2021)Lin, Sun, Dhingra, Zaheer, Ren, and
  Cohen}]{differentiable}
Bill~Yuchen Lin, Haitian Sun, Bhuwan Dhingra, Manzil Zaheer, Xiang Ren, and
  William Cohen. 2021.
\newblock Differentiable open-ended commonsense reasoning.
\newblock In \emph{Proceedings of the 2021 Conference of the North American
  Chapter of the Association for Computational Linguistics: Human Language
  Technologies}, pages 4611--4625.

\bibitem[{Muzellec and Cuturi(2018)}]{elliiptical}
Boris Muzellec and Marco Cuturi. 2018.
\newblock Generalizing point embeddings using the wasserstein space of
  elliptical distributions.
\newblock NIPS'18, page 10258–10269.

\bibitem[{Nickel and Kiela(2017)}]{nickel2017poincare}
Maximillian Nickel and Douwe Kiela. 2017.
\newblock Poincar{\'e} embeddings for learning hierarchical representations.
\newblock \emph{Advances in neural information processing systems}, 30.

\bibitem[{Onoe et~al.(2021)Onoe, Boratko, McCallum, and Durrett}]{box4et}
Yasumasa Onoe, Michael Boratko, Andrew McCallum, and Greg Durrett. 2021.
\newblock Modeling fine-grained entity types with box embeddings.
\newblock In \emph{ACL}.

\bibitem[{Ren et~al.(2020)Ren, Hu, and Leskovec}]{query2box}
Hongyu Ren, Weihua Hu, and Jure Leskovec. 2020.
\newblock \href {https://openreview.net/forum?id=BJgr4kSFDS} {Query2box:
  Reasoning over knowledge graphs in vector space using box embeddings}.
\newblock In \emph{International Conference on Learning Representations}.

\bibitem[{Sun et~al.(2019)Sun, Deng, Nie, and Tang}]{rotatE}
Zhiqing Sun, Zhi-Hong Deng, Jian-Yun Nie, and Jian Tang. 2019.
\newblock Rotate: Knowledge graph embedding by relational rotation in complex
  space.
\newblock In \emph{International Conference on Learning Representations}.

\bibitem[{Trouillon et~al.(2016)Trouillon, Welbl, Riedel, Gaussier, and
  Bouchard}]{complexE}
Th\'{e}o Trouillon, Johannes Welbl, Sebastian Riedel, \'{E}ric Gaussier, and
  Guillaume Bouchard. 2016.
\newblock Complex embeddings for simple link prediction.
\newblock In \emph{Proceedings of the 33rd International Conference on
  International Conference on Machine Learning - Volume 48}, ICML'16, page
  2071–2080. JMLR.org.

\bibitem[{Vilnis et~al.(2018)Vilnis, Li, Murty, and McCallum}]{box}
Luke Vilnis, Xiang~Lorraine Li, Shikhar Murty, and Andrew McCallum. 2018.
\newblock Probabilistic embedding of knowledge graphs with box lattice
  measures.
\newblock In \emph{ACL}.

\bibitem[{Wang et~al.(2021{\natexlab{a}})Wang, Huang, Liu, Shao, Liu, Li, Wang,
  Sun, Yao, and Abdelzaher}]{DyDiff-VAE}
Ruijie Wang, Zijie Huang, Shengzhong Liu, Huajie Shao, Dongxin Liu, Jinyang Li,
  Tianshi Wang, Dachun Sun, Shuochao Yao, and Tarek Abdelzaher.
  2021{\natexlab{a}}.
\newblock Dydiff-vae: A dynamic variational framework for information diffusion
  prediction.
\newblock In \emph{SIGIR'21}.

\bibitem[{Wang et~al.(2021{\natexlab{b}})Wang, Wei, Nogueira~dos Santos, Wang,
  Nallapati, Arnold, Xiang, Yu, and Cruz}]{hyperbolic_kg}
Shen Wang, Xiaokai Wei, Cicero~Nogueira Nogueira~dos Santos, Zhiguo Wang,
  Ramesh Nallapati, Andrew Arnold, Bing Xiang, Philip~S. Yu, and Isabel~F.
  Cruz. 2021{\natexlab{b}}.
\newblock Mixed-curvature multi-relational graph neural network for knowledge
  graph completion.
\newblock In \emph{Proceedings of the Web Conference 2021}, WWW '21, page
  1761–1771.

\bibitem[{Wang et~al.(2021{\natexlab{c}})Wang, Wei, Nogueira~dos Santos, Wang,
  Nallapati, Arnold, Xiang, Yu, and Cruz}]{m2gnn}
Shen Wang, Xiaokai Wei, Cicero~Nogueira Nogueira~dos Santos, Zhiguo Wang,
  Ramesh Nallapati, Andrew Arnold, Bing Xiang, Philip~S Yu, and Isabel~F Cruz.
  2021{\natexlab{c}}.
\newblock Mixed-curvature multi-relational graph neural network for knowledge
  graph completion.
\newblock In \emph{Proceedings of the Web Conference 2021}, pages 1761--1771.

\bibitem[{Wang et~al.(2014)Wang, Zhang, Feng, and Chen}]{TransH}
Zhen Wang, Jianwen Zhang, Jianlin Feng, and Zheng Chen. 2014.
\newblock Knowledge graph embedding by translating on hyperplanes.
\newblock \emph{Proceedings of the 28th AAAI Conference on Artificial
  Intelligence}, 28.

\bibitem[{Yang et~al.(2015)Yang, Yih, He, Gao, and Deng}]{dismult}
Bishan Yang, Wen-tau Yih, Xiaodong He, Jianfeng Gao, and Li~Deng. 2015.
\newblock Embedding entities and relations for learning and inference in
  knowledge bases.
\newblock In \emph{International Conference on Learning Representations
  (ICLR)}.

\bibitem[{Yao et~al.(2020)Yao, Mao, and Luo}]{KGbert}
Liang Yao, Chengsheng Mao, and Yuan Luo. 2020.
\newblock Kg-bert: Bert for knowledge graph completion.
\newblock \emph{Proceedings of the Thirty-Fourth AAAI Conference on Artificial
  Intelligence}.

\bibitem[{Yasunaga et~al.(2021)Yasunaga, Ren, Bosselut, Liang, and
  Leskovec}]{qagnn}
Michihiro Yasunaga, Hongyu Ren, Antoine Bosselut, Percy Liang, and Jure
  Leskovec. 2021.
\newblock Qa-gnn: Reasoning with language models and knowledge graphs for
  question answering.
\newblock In \emph{North American Chapter of the Association for Computational
  Linguistics (NAACL)}.

\bibitem[{Zhang et~al.(2020)Zhang, Zhuang, Zhu, Shi, Xiong, and He}]{aaai_gnn}
Zhao Zhang, Fuzhen Zhuang, Hengshu Zhu, Zhiping Shi, Hui Xiong, and Qing He.
  2020.
\newblock Relational graph neural network with hierarchical attention for
  knowledge graph completion.
\newblock \emph{Proceedings of the AAAI Conference on Artificial Intelligence},
  pages 9612--9619.

\end{thebibliography}

\newpage

\appendix
\begin{figure*}[htbp]
    \centering
   \includegraphics[width=\linewidth]{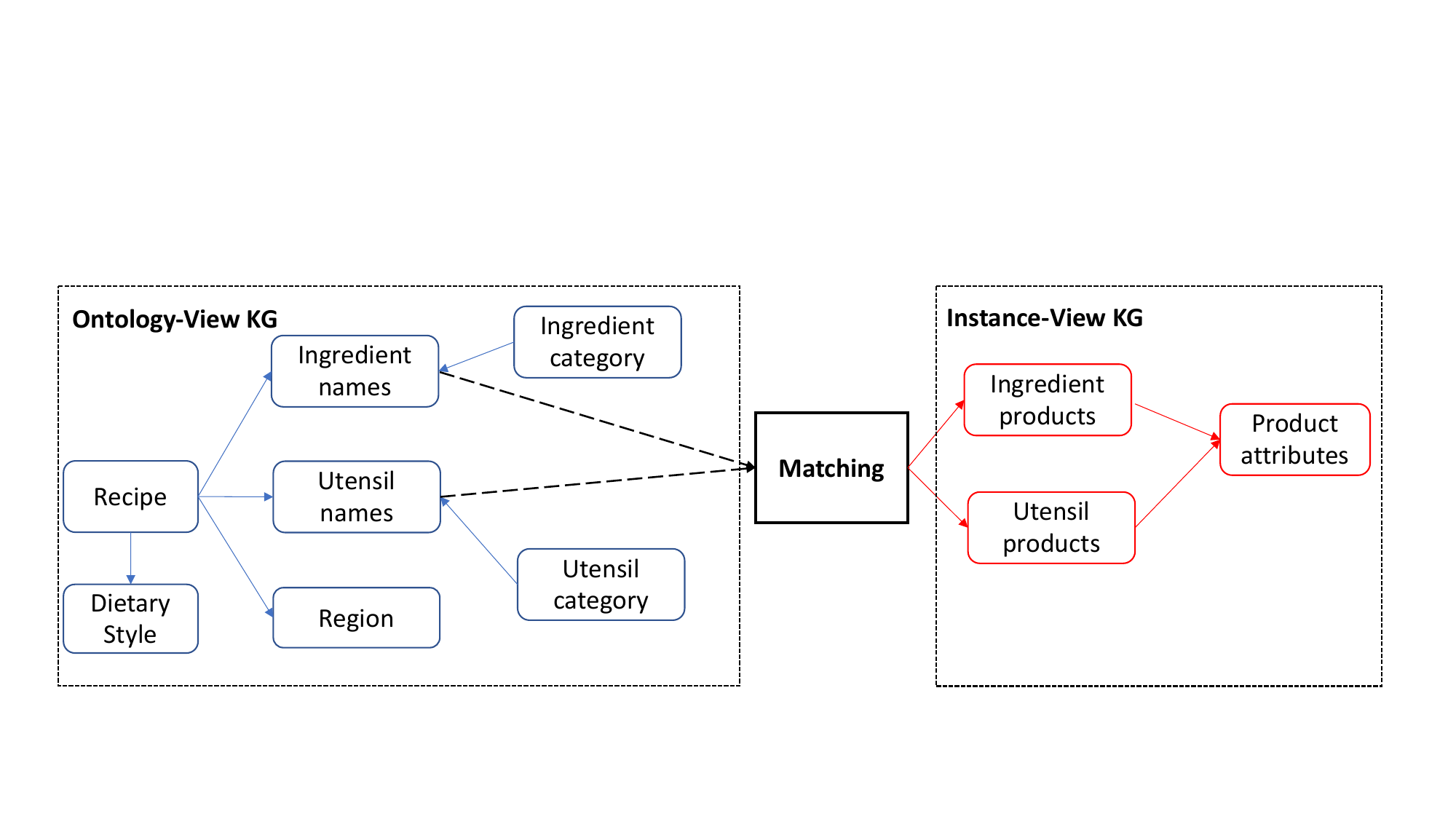}
  \caption{Recipe Dataset Structure}
  \label{fig:recipe}
\end{figure*}   
\section{Data Construction}~\label{appendix1}
\label{data construct}
We introduce the data generation process of the Recipe dataset. The overall data structure is depicted in Fig~\ref{fig:recipe}. For the ontology-view KG, concepts consist of recipes, ingredients, and utensils for cooking them and their associated types. Dietary styles and regions where each recipe originated from are also considered concepts. We obtain these high-level recipe-related concepts from  recipeDB~\footnote{https://cosylab.iiitd.edu.in/recipedb/} with some data cleaning procedures. Triples in the ontology-view KG describe the connectivity among these concepts such as (\textit{Kungpao Chicken}, \textit{use$\_$ingredient}, \textit{Chicken Breast}), (\textit{Salt}, \textit{is$\_$of$\_$ingredient$\_$type}, \textit{Additive}). 

For the instance-view KG, entities are real ingredient and utensil products sold on Amazon.com by searching their corresponding concept names (ingredient and utensil names) as keywords. For each ingredient and utensil name, we select the top 10 matched products as entities in the instance-view KG. Afterward, we manually select 8 representative attributes such as flavor and brand, that are commonly shared by these products as additional entities in the instance-view KG, to provide more knowledge for the embedding learning process. The triples in the instance-view KG describe facts connecting a product to its attributes.

\section{Implementation Details.}

We use Adam~\cite{adam} as the optimizer to train our model and use TransE~\cite{transE} as the KGE model for learning the instance-view KG, whose margin $\gamma^{\text{KG}}$ is set to be 0.3. The margin for cross-view module $\gamma^{\text{Cross}}$ is set to be 0.15.  We set the latent dimensions for instance-view KG as 256, and the box dimensions for ontology-view KG as 512.  We use a batch size of 128 and a learning rate $lr\!=\!0.005$ during training.

Instead of directly optimizing $\mathcal{J}$ in Eqn~\ref{eq:loss_all}, we alternately update $\mathcal{J}^{\mathcal{G}^{\mathcal {O }}}$, $\mathcal{J}^{\mathcal{G}^{\mathcal {I}}}$ and $\mathcal{J}^{\text {Cross }}$ with different learning rate. Specifically, in our implementation, we optimize with $\theta_{new}\leftarrow \theta_{old} - \eta \nabla \mathcal{J}^{\mathcal{G}^{\mathcal {O }}}$, $\theta_{new}\leftarrow \theta_{old} - (\lambda_1\eta) \nabla \mathcal{J}^{\mathcal{G}^{\mathcal {I }}}$, $\theta_{new}\leftarrow \theta_{old} - (\lambda_2\eta) \nabla \mathcal{J}^{\text {cross }}$ in consecutive steps within one epoch, where $\theta_{new}$ denotes our model parameters. The procedure is depicted in Algorithm~\ref{alg1}.

\begin{algorithm}[htbp]
\caption{Concept2Box training procedure.}
\KwIn{A two-view KG $\mathcal{G}$.
\label{alg1}
\noindent\KwOut{Model parameters $\theta$.}
\While{model not converged}{
//\textit{For the KGC loss of $\mathcal{G}^{\mathcal{O}}$}:\\
 Optimize the box-based KGC loss in Eqn~\ref{eq:loss_O}:\\
 $\theta_{new}\leftarrow \theta_{old} - \eta \nabla \mathcal{J}^{\mathcal{G}^{\mathcal {O }}}$\\
//\textit{For the KGC loss of $\mathcal{G}^{\mathcal{I}}$}:\\
Optimize the vector-based KGC loss in Eqn~\ref{eq:loss_I}:\\
 $\theta_{new}\leftarrow \theta_{old} - (\lambda_1\eta) \nabla \mathcal{J}^{\mathcal{G}^{\mathcal {I}}}$\\
//\textit{For the cross-view loss}:\\
Optimize with the cross-view loss in Eqn~\ref{eq:loss_cross}:\\
$\theta_{new}\leftarrow \theta_{old} - (\lambda_2\eta) \nabla \mathcal{J}^{\text{cross}}$\\
    }
}

\end{algorithm}

\end{document}